\documentclass{article}

\usepackage[nonatbib, final]{nips_2016}

\usepackage[utf8]{inputenc} 
\usepackage[T1]{fontenc}    
\usepackage{hyperref}       
\usepackage{url}            
\usepackage{booktabs}       
\usepackage{amsfonts}       
\usepackage{nicefrac}       
\usepackage{microtype}      
\usepackage{amsmath}
\usepackage{amssymb}
\usepackage{graphicx}

\title{metricDTW: local distance metric learning in Dynamic Time Warping}

\author{
  Jiaping Zhao, Zerong Xi, Laurent Itti \\
  Department of Computer Science\\
  University of Southern California\\
  \texttt{\{jiapingz,zxi,itti\}@usc.edu} \\
}

\begin{document}

\maketitle

\begin{abstract}
We propose to learn multiple local Mahalanobis distance metrics to
perform k-nearest neighbor (kNN) classification of temporal sequences.
Temporal sequences are first aligned by dynamic time warping (DTW);
given the alignment path, similarity between two sequences is measured
by the DTW distance, which is computed as the accumulated distance
between matched temporal point pairs along the alignment path. Traditionally,
Euclidean metric is used for distance computation between matched
pairs, which ignores the data regularities and might not be optimal
for applications at hand. Here we propose to learn multiple Mahalanobis
metrics, such that DTW distance becomes the sum of Mahalanobis distances.
We adapt the large margin nearest neighbor (LMNN) framework to our
case, and formulate multiple metric learning as a linear programming
problem. Extensive sequence classification results show that our proposed
multiple metrics learning approach is effective, insensitive to the
preceding alignment qualities, and reaches the state-of-the-art performances
on UCR time series datasets.
\end{abstract}

\section{Introduction}

Dynamic time warping (DTW) is an algorithm to align temporal sequences
and measure their similarities. DTW has been widely used in speech
recognition \cite{rabiner1993fundamentals}, human motion synthesis
\cite{hsu2005style}, human activity recognition \cite{kulkarni2014continuous}
and time series classification \cite{UCRArchive}. DTW allows temporal
sequences to be locally shifted, contracted and stretched, and it
calculates a global optimal alignment path between two given sequences
under certain restrictions. Therefore, the similarity between two
sequences calculated under the optimal alignment is independent of,
to some extent, non-linear variations in the time dimension. The similarity
is often quantified by the DTW distance, which is the sum of point-wise
distances along the alignment path, i.e., $D(\mathcal{P},\mathcal{Q})=\Sigma_{(i,j)\in\textbf{p}}d(i,j)$,
where $\textbf{p}$ is the alignment path between two sequences $\mathcal{P}$
and $\mathcal{Q}$, $(i,j)$ is a pair of matched points on the alignment
path and $d(i,j)$ is the distance (affinity) between $i$ and $j$.
The most widely used point-wise distance $d(i,j)$ is the (squared)
Euclidean distance.

Since DTW distance naturally measures the similarity between time
series, it is widely used for time series classification. There is
increasing acceptance that the nearest neighbor classifier with the
DTW distance as the similarity measure (1NN-DTW) is the choice for
most time series classification problems and very hard to beat \cite{petitjean2014dynamic,wang2013experimental,bagnall2014experimental,rakthanmanon2012searching}.
Although 1NN-DTW is competitive and hard to beat, to the best of our
knowledge, the DTW distance is often computed as the sum of point-wise
(squared) Euclidean distances along the matching path, i.e., $D(\mathcal{P},\mathcal{Q})=\Sigma_{(i,j)\in\textbf{p}}d(i,j)$,
where $d(i,j)=\parallel i-j\parallel^{2}$ is the (squared) Euclidean
distance between the matched points $i$ and $j$. Nevertheless, the
performance of kNN significantly depends on the used similarity measures.
Although Euclidean distance is simple and sometimes effective, but
it is agnostic of domain knowledge and data regularities. Extensive
researches have shown that kNN performances can be greatly improved
by learning a proper distance metric (e.g., Mahalanobis distance)
from labels examples \cite{weinberger2009distance,bellet2013survey}.
This motives us to learn local distance metrics and calculate DTW
distance as the sum of point-wise learned distances, i.e., $\hat{D}(\mathcal{P},\mathcal{Q})=\Sigma_{(i,j)\in\textbf{p}}\hat{d}(i,j)$,
where $\hat{d}(i,j)=(i-j)^{T}\mathcal{M}_{ij}(i-j)$, $\mathcal{M}_{ij}$
is a positive semidefinite matrix to be learned and $\hat{d}(i,j)$
is the squared Mahalanobis distance. In the paper, instead of learning
one uniform distance metric, we partition the feature space, and learn
individual metrics within and between subspaces. When using DTW distance
calculated under the learned metrics as the similarity measure, 1NN
classifier has the potential to obtain improved performances.

We closely follow Large Margin Nearest Neighbor (LMNN) \cite{weinberger2009distance}
to formulate local metric learning in DTW. In \cite{weinberger2009distance},
the Mahalanobis metric is learned with the goal that the k-nearest
neighbors always belong to the same class while examples from different
classes are separated by a large margin. Mathematically, the authors
formulate the metric learning as a semidefinite programming problem.
In our case, we use the same max margin framework, with the only difference
that: examples in \cite{weinberger2009distance} are feature points
in some fixed-dimension space, and distances between examples are
squared Mahalanobis distances, while in our case, examples are temporal
sequences, and distances between examples are DTW distances. We term
the local metric learning in DTW as metricDTW. We have to emphasize
that although the learned local distances are metric distances, the
DTW distance under those metrics is generally not a metric distance
since the triangle inequality does not hold.

\begin{figure*}
\begin{centering}
\includegraphics[width=1.0\textwidth]{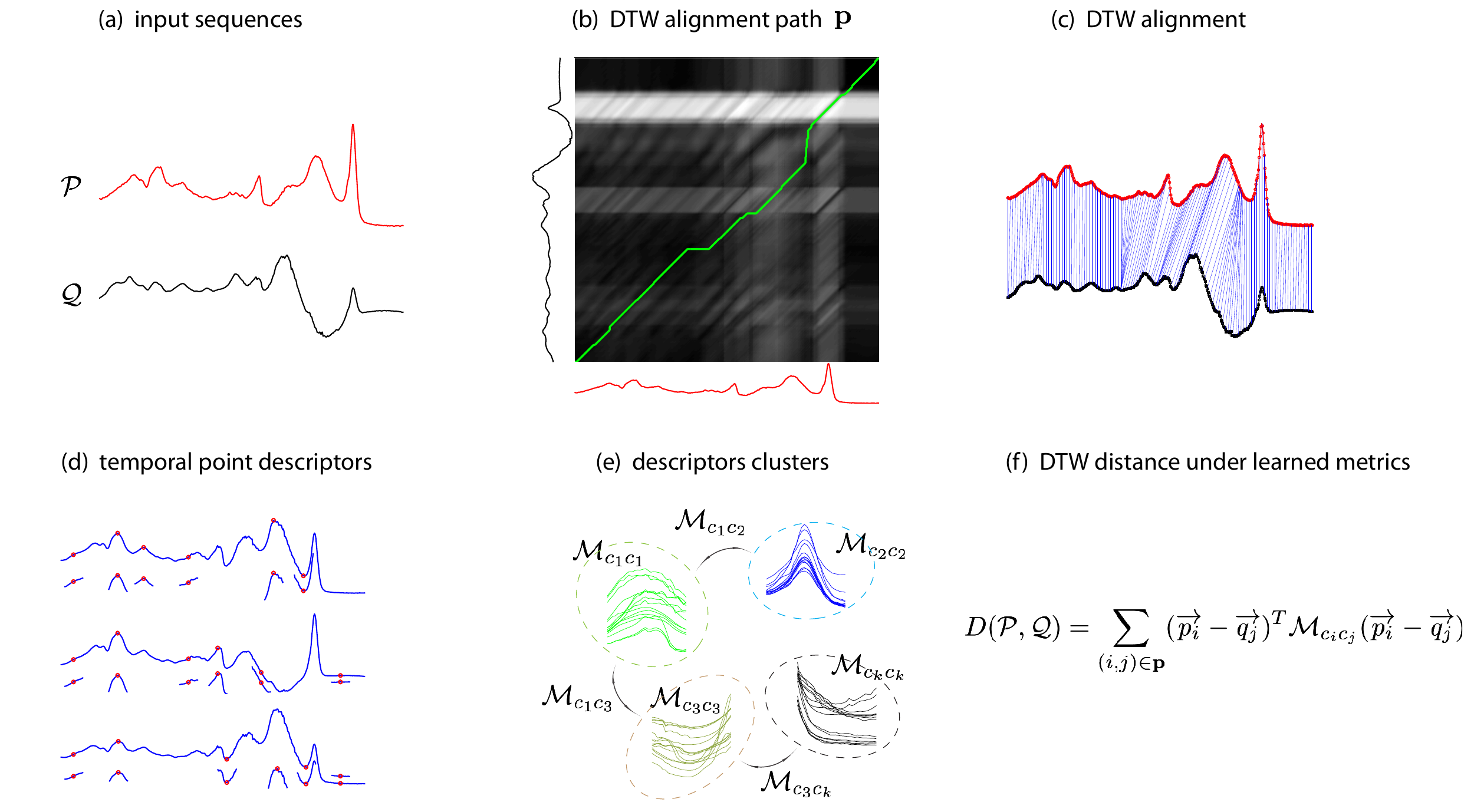} 
\par\end{centering}

\centering{}\caption{\label{fig:flowchart}Multiple local distance metrics learning in
DTW. In the paper, we propose to learn multiple local Mahalanobis
distance metrics to perform k-nearest neighbor (kNN) classification
of temporal sequences. The similarity between two given sequences
is measured by their DTW distance (f), which is calculated as the
accumulated Mahalanobis distances between the matched point pairs
along the alignment path. As a preceding step for our metric learning
algorithm, DTW is used to compute the alignment path (b,c). Different
from the tradition, we compute the distance between a matched point
pair by the distance between their descriptors (d), and if we further
partition the descriptor space into $k$ clusters and define an individual
metric within each cluster and between any two clusters (e), then
the DTW distance will take the form as in (f). We adapt LMNN \cite{weinberger2009distance}
to formulate our multiple metric learning in DTW.}
\end{figure*}

Before computing the DTW distance, we have to align given sequences
to obtain the alignment path, along which the DTW distance is defined.
In our work, we do not aim to learn to align sequences, instead, we
use existing DTW techniques to align sequences first and treat these
alignment paths as known. Therefore, the metric learning in DTW is
independent of the preceding alignment process, in principle, any
sequence alignment algorithms can be used before metric learning.

In the paper, different from the tradition, we compute the distance
between a matched point pair by the distance between their descriptors.
The descriptor of a temporal point is a representation of the subsequence
centering on that point, and it represents the structural information
around that point (see Fig. \ref{fig:Descriptor-of-temporal}). In
this way, DTW distance is computed as the accumulated descriptor distances
along the alignment path. In our case, descriptors are further k-means
clustered into groups, then multiple local distance metrics are learned
within individual clusters and between any two clusters, such that
DTW distances calculated under the learned metrics make kNN neighbors
of temporal sequences always come from the same class, while sequences
from different classes are separated by a large margin. Our proposed
local metrics learning framework is depicted in Fig. \ref{fig:flowchart}.

Our multiple metric learning ends up learning multiple Mahalanobis
distance matrices, some magnifying the Euclidean distances between
subsequences, while other shrinking the original Euclidean distances.
This is equivalent to learn the importance of subsequences automatically,
e.g., certain shaped subsequences are discriminative for classification,
and their subtle differences may be magnified by the corresponding
learned metrics, while certain shaped subsequences are less class-membership
defining, and their big differences might be suppressed after metric
learning. In this perspective, our local metric learning framework
is essentially learning the importance of different subsequences in
an automatic and principled way. 

We extensively test the performance of metricDTW for time series classification
on 70 UCR time series datasets \cite{UCRArchive}, and experimental
results show that (1) the learned local metrics, compared with the
default Euclidean metric, improve the 1NN classification accuracies
significantly; (2) given alignment paths of different qualities, the
subsequent metric learning consistently boosts classification accuracies
significantly, showing that the proposed metric learning approach
is invariant to the preceding alignment step; (3) our metric learning
algorithm outperforms the state-of-the-art time series classification
algorithm (1NN-DTW) significantly on UCR datasets, therefore, we set
a new record for future time series classification comparison.

\section{Related work}

As mentioned, our local metric learning framework is essentially learning
the importance of different subsequences in an automatic and principled
way. There are several prior works focusing on mining representative
and discriminative subsequences (image patches) from temporal sequences
(images).

Time series shapelet is introduced in \cite{ye2009time}, and it is
a time series subsequence (patterns) which is discriminative of class-membership.
The authors propose to enumerate all possible candidate subsequences,
evaluate their qualities using information gain, and build a decision
tree classifier out of the top ranked shapelets. Mining shapelets
in their case is to search for more important subsequences, while
disregarding less important subsequences. In the vision community,
there are several related works \cite{singh2012unsupervised,doersch2012makes,doersch2013mid},
all of which are devoted to discovering mid-level visual patches from
images. Mid-level visual patch is conceptually similar to shapelet
in time series, and it a image patch which is both representative
and discriminative for scene categories. They \cite{singh2012unsupervised,doersch2012makes}
pose the discriminative patch search procedure as a discriminative
clustering process, in which they selectively choose important patches
but discarding other common patches. We are different from above work
in that, we never have to greedily select important subsequences,
instead, we take all subsequences into account and automatically learn
their importance through metric learning. 

Our work is most similar to and largely inspired by LMNN \cite{weinberger2009distance}.
In \cite{weinberger2009distance}, Weinbergre and Saul extend LMNN
to learn multiple local distance metrics, which is exploited in our
work as well. However, we are still sufficiently different: first
the labeled examples in our case are temporal sequences; second, the
DTW distance between two examples is jointly defined by multiple metrics,
while in \cite{weinberger2009distance}, distance between two examples
are determined by a single metric. In \cite{garreau2014metric}, Garreau
et al propose to learn a Mahalanobis distance metric to perform DTW
sequence alignment. First they need ground-truth alignments, which
is not required in our case, and second they focus on alignment, instead
of kNN classification.

\section{Local distance metric learning in DTW}

As mentioned above, local metric learning needs sequence alignments
as inputs. While in most scenarios, ground-truth sequence to sequence
alignments are expensive or impossible to label, in experiments, we
use DTW to align sequences first, and use the computed alignments
for the subsequent metric learning. In this section, we first briefly
review the DTW algorithm for sequence alignment, and then introduce
our multiple local metric learning algorithm for time series classification.

\subsection{Dynamic Time Warping}

DTW is an algorithm to align temporal sequences under certain restrictions.
Given two sequences $\mathcal{P}$ and $\mathcal{Q}$ of possible
different lengths $\mathcal{L_{P}}$ and $\mathcal{L_{Q}}$, namely
$\mathcal{P}=(p_{1},p_{2},...,p_{\mathcal{L_{P}}})^{T}$ and $\mathcal{Q}=(q_{1},q_{2},...,q_{\mathcal{L_{Q}}})^{T}$,
and let $d(\mathcal{P},\mathcal{Q})\in\mathcal{R^{L_{P}\times L_{Q}}}$
be the pairwise distance matrix, where $d(i,j)$ is the distance between
points $p_{i}$ and $p_{j}$. One widely used distance measure is
the squared Euclidean distance, i.e., $d(i,j)=\parallel p_{i}-q_{j}\parallel_{2}^{2}$.
The goal of temporal alignment between $\mathcal{P}$ and $\mathcal{Q}$
is to find two sequences of indices $\alpha$ and $\beta$ of the
same length $l$, which match index $\alpha(i)$ in the time series
$\mathcal{P}$ to index $\beta(i)$ in the time series $\mathcal{Q}$,
such that the total cost along the matching path $\sum_{i=1}^{l}d(\alpha(i),\beta(i))$
is minimized. The alignment path $\textbf{p}=(\alpha,\beta)$ is constrained
to satisfies boundary, monotonicity and step-pattern conditions \cite{sakoe1978dynamic,keogh2005exact,garreau2014metric}:

\vspace{-6pt}

\begin{equation}
\begin{cases}
\begin{array}{l}
\alpha(1)=\beta(1)=1,\;\alpha(l)=\mathcal{L_{P}},\beta(l)=\mathcal{L_{Q}}\\
\alpha(1)\le\alpha(2)\le...\le\alpha(l),\:\beta(1)\le\beta(2)\le...\le\beta(l)\\
\left(\alpha(i+1),\beta(i+1))-(\alpha(i),\beta(i)\right)\in\left\{ (1,0),(1,1),(0,1)\right\} 
\end{array}\end{cases}\label{eq:DTWconditions}
\end{equation}

Searching for an optimal alignment path $\textbf{p}$ under the above
restrictions is equivalent to solve the following recursive formula:

\begin{equation}
\textbf{D}(i,j)=d(i,j)+\min\{\textbf{D}(i-1,j-1),\,\textbf{D}(i,j-1),\,\mathcal{\textbf{D}}(i-1,j)\}\label{eq:DTW}
\end{equation}

where $\textbf{D}(i,j)$ is the accumulated distance from the matched
point-pair $(p_{1},q_{1})$ to the matched point-pair $(p_{i},q_{j})$
along the alignment path, and $d(i,j)$ is the distance between points
$p_{i}$ and $p_{j}$. In all the following alignment experiments,
we use the squared Euclidean distance to compute $d(i,j)$. The above
formula is a typical dynamic programming recursion, and can be solved
efficiently in $\mathcal{O}(\mbox{\ensuremath{\mathcal{L_{P}}\times}\ensuremath{\ensuremath{\mathcal{L_{Q}}}}})$
time by a dp algorithm \cite{EllisDTW}. The alignment path $\textbf{p}$
is obtained by back-tracking. Various temporal window constraints
\cite{sakoe1978dynamic} can be enforced and we could use more complicated
step patterns, such as ``asymmetric'' and ``rabinerJuang'' \cite{rabiner1993fundamentals,giorgino2009computing},
but here we consider DTW without warping window constraints and taking
moving patterns as defined in (\ref{eq:DTWconditions}).

\subsection{Local distance metric learning\label{sub:Local-distance-metric}}

\begin{figure*}
\begin{centering}
\includegraphics[width=0.5\textwidth]{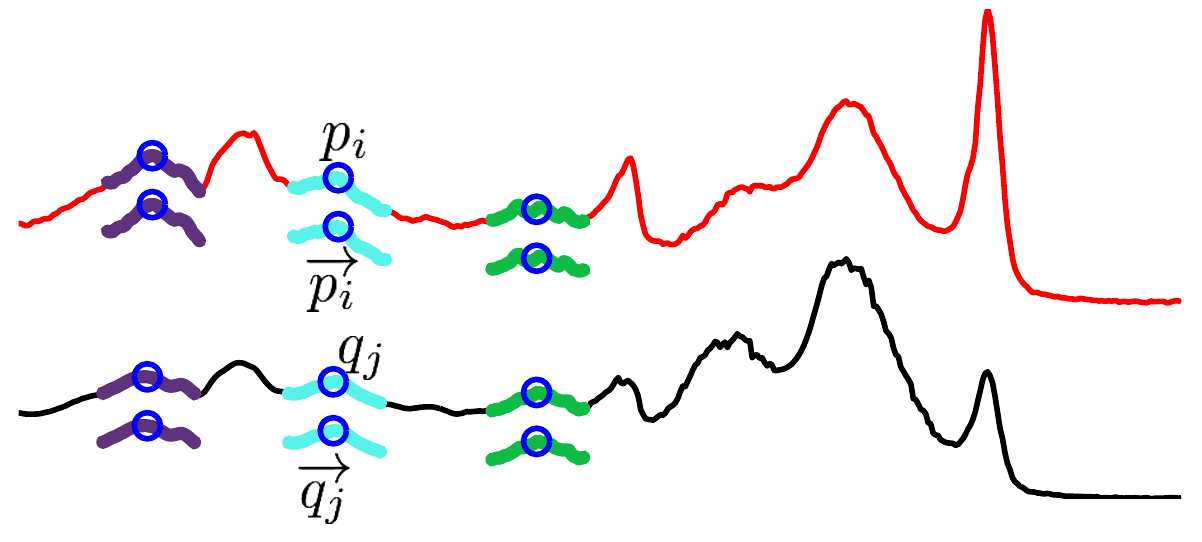} 
\par\end{centering}

\centering{}\caption{\label{fig:Descriptor-of-temporal}Descriptor of temporal point. As
shown, $p_{i}$ and $q_{j}$ are temporal points from sequences, and
the descriptor of a temporal point is defined to the representation
of the subsequence centering on that point, e.g., the bold cyan subsequence
($\protect\overrightarrow{p}$ ) around $p_{i}$ is its descriptor,
and any representation of $\protect\overrightarrow{p}$ is called
the descriptor of $p_{i}$ as well, like HOG-1D and derivative sequence.}
\end{figure*}

After obtaining the alignment path $\textbf{p}$ by DTW , we can compute
DTW distance between $\mathcal{P}$ and $\mathcal{Q}$ in two ways:
(1) directly return DTW distance as the accumulated distances between
matched pairs along $\textbf{p}$, i.e., $\sum_{(i,j)\in\textbf{p}}d(p_{i},q_{j})$;
(2) to measure the distance between a matched pair $(p_{i},q_{j})$,
we could use the distance between their descriptors, i.e., $d(\overrightarrow{p_{i}},\overrightarrow{q_{j}})$,
where $\overrightarrow{p_{i}}$ and $\overrightarrow{q_{j}}$ are
descriptors of points $p_{i}$ and $q_{j}$ respectively. In this
way, DTW distance between $\mathcal{P}$ and $\mathcal{Q}$ is calculated
as the accumulated descriptor distances along $\textbf{p}$, i.e.,
$\sum_{(i,j)\in\textbf{p}}d(\overrightarrow{p_{i}},\overrightarrow{q_{j}})$.
Here, the descriptor at some point is a feature vector representation
of the subsequence centering at that point, and the descriptor is
supposed to capture the neighborhood shape information around the
temporal point (see Fig. \ref{fig:Descriptor-of-temporal} for the
illustration of descriptors). Using their descriptor distance to measure
two point similarity (distance) makes much sense since two point similarity
is usually better represented by their neighbor structural similarity,
instead of by their single point to point distance. 

In following experiments, we always adopt the second way to define
the DTW distance, and we use three shape descriptors, namely the raw-subsequence,
HOG-1D \cite{zhaodecomposing} and the gradient sequence \cite{keogh2001derivative}.

If the squared Euclidean distance is used, then DTW distance is calculated
as $D(\mathcal{P},\mathcal{Q})=\sum_{(i,j)\in\textbf{p}}\parallel\overrightarrow{p_{i}}-\overrightarrow{q_{j}}\parallel^{2}$,
which is essentially a equally weighted sum of distances between descriptors
(subsequences), however, as shown in \cite{ye2009time}, some subsequences
are more class-membership predictive, while others are less discriminative.
Therefore, it makes more sense, if we calculate the DTW distance as
a weighed sum of distances between subsequences, i.e., $D(\mathcal{P},\mathcal{Q})=\sum_{(i,j)\in\textbf{p}}\omega_{ij}\parallel\overrightarrow{p_{i}}-\overrightarrow{q_{j}}\parallel^{2}$,
where $\omega_{ij}$ is the weight between two subsequences $\overrightarrow{p_{i}}$
and $\overrightarrow{q_{i}}$ and indicates the importance of subsequences
$\overrightarrow{p_{i}}$ and $\overrightarrow{q_{i}}$ for classification.
If we make further generalization, the DTW distance can be calculated
as the sum of squared Mahalanobis distances between subsequences,
i.e., $D(\mathcal{P},\mathcal{Q})=\sum_{(i,j)\in\textbf{p}}(\overrightarrow{p_{i}}-\overrightarrow{q_{j}})^{T}\mathcal{M}_{c_{i}c_{j}}(\overrightarrow{p_{i}}-\overrightarrow{q_{j}})$,
where $\mathcal{M}_{c_{i}c_{j}}$ is a positive semidefinite Mahalanobis
matrix to be learned from the labeled data. Note that, instead of
learning a global metric matrix, we learn multiple local metric matrices
simultaneously. The intuition behind is that differently-shaped subsequences
have different importance for classification, therefore, their between-distances
should be computed under different metrics. In experiments, we first
k-means partition the descriptors from all training sequences into
$k$ clusters, and then learn Mahalanobis distance metrics within
individual clusters and between any two different clusters. Let$\mathcal{M}_{c_{i}c_{i}}$,
$\mathcal{M}_{c_{i}c_{j}}$ denote the metrics within the cluster
$c_{i}$ and between two cluster $c_{i}$ and $c_{j}$ respectively,
and then the distance between any two descriptors $\overrightarrow{p_{i}}$
and $\overrightarrow{q_{i}}$ is $(\overrightarrow{p_{i}}-\overrightarrow{q_{j}})^{T}\mathcal{M}_{c_{i}c_{j}}(\overrightarrow{p_{i}}-\overrightarrow{q_{j}})$,
where $c_{i}$ and $c_{j}$ are clusters $\overrightarrow{p_{i}}$
and $\overrightarrow{q_{i}}$ belong to respectively.

In order to learn these local metrics from labeled sequence data,
we follow LMNN \cite{weinberger2009distance} closely and pose our
problem as a max margin problem: the local Mahalanobis metrics are
trained such that the k-nearest neighbors of any sequence always belong
to the same class while sequences of different classes are separated
by a large margin. We use the exact notations in LMNN, and the only
place to change is to replace the squared Mahalanobis point-to-point
distance in \cite{weinberger2009distance} by the DTW distance. The
adapted LMNN is as follows:

\vspace{-10pt}
\begin{equation}
\begin{array}{l}
Minimize:\;(1-\mu)\Sigma_{i,j\rightsquigarrow i}D(x_{i},x_{j})+\mu\Sigma_{j\rightsquigarrow i,l}(1-y_{il})\xi_{ijl}\\
Subject\: to:\\
\qquad(1)\: D(x_{i},x_{l})-D(x_{i},x_{j})\geq1-\xi_{ijl}\\
\qquad(2)\:\xi_{ijl}\geq0\\
\qquad(3)\:\mathcal{M}_{c_{i}c_{j}}\equiv\mathcal{M}_{c_{j}c_{i}},\:\mathcal{M}_{c_{i}c_{j}}\succeq0,\quad c_{i},c_{j}\in\{1,2,...,k\}
\end{array}\label{eq:LMNN-dtw}
\end{equation}

Note that we enforce the learned matrices between two clusters $c_{i}$
and $c_{j}$ to be the same, i.e., $\mathcal{M}_{c_{i}c_{j}}\equiv\mathcal{M}_{c_{j}c_{i}}$,
which makes distance mapping between $c_{i}$ and $c_{j}$ be a metric.
We refer readers to \cite{weinberger2009distance} for notation meanings.
In our experiments, we further simplify the form of Mahalanobis matrices,
and constrain them to be not only diagonal but also with single repeated
element on the diagonal, i.e., $\mathcal{M}_{c_{i}c_{j}}=\omega_{c_{i}c_{j}}\cdot\textbf{I}$.
Under this simplification, learning a Mahalanobis matrix reduces to
learning a scalar, resulting in $(k^{2}+k)/2$ unknown scalars to
learn. Under the reduction, the original semidefinite programming
(\ref{eq:LMNN-dtw}) reduces to a linear programming. In experiments,
the balancing factor $\mu$ is tuned by cross-validation.

\section{Experiments}

In this section, we evaluate the performances of the proposed local
metric learning method for time series classification on 70 UCR datasets
\cite{UCRArchive}, which provide their standard training/test partitions
for performance evaluation.

We empirically show below that: (1) whether multiple local metric
learning boosts time series classification accuracies of 1NN classifier;
(2) how the quality of the preceding alignments affects the subsequent
metric learning performances; (3) the influence of hyper-parameter
settings on the metric learning performances.

\subsection{Experimental settings}

Sequence alignment: when running DTW (\ref{eq:DTW}) to align sequences,
we use the default squared Euclidean distances to compute point-to-point
distance. The alignment paths are the inputs of the subsequent metric
learning step.

Temporal point descriptors: the descriptor at a temporal point is
used to represent its neighborhood structures. When computing the
point to point distance in DTW alignment ($d(i,j)$ in (\ref{eq:DTW})),
we compute the distance between their descriptors and use it as the
distance between the original temporal points. Obviously, the searched
optimal alignment path by DTW (\ref{eq:DTW}) depends on the used
descriptor for point-to-point distance computations. Descriptors are
used in the subsequent metric learning as well to define the DTW distance
(see Sec. \ref{sub:Local-distance-metric}).

In experiments, we use three subsequence descriptors, including raw-subsequence,
HOG-1D \cite{zhaodecomposing} and the derivative sequence \cite{keogh2001derivative}.
(1) The raw-subsequences taken from temporal points are fixed to be
of length 30; (2) HOG-1D is a representation of the raw subsequence,
and we use two non-overlapping intervals, use 8 bins and set $\sigma=0.1,$
resulting in a 16D HOG-1D descriptor; (3) the derivative descriptor
is simply the first order derivative sequence of the raw subsequence.
We follow \cite{keogh2001derivative} exactly to compute derivative
at each point, and the derivative descriptor is 30D by definition.

Metric learning: k in kNN is set to 3. For each training time series,
we compute its 3 nearest neighbors of the same class based on the
DTW distances, which is computed under the default Euclidean metric.
We set k in k-means to be 5, partition the training descriptors into
5 clusters and local distance metrics are defined within and between
these 5 clusters. The linear program (\ref{eq:LMNN-dtw}) is solved
by the CVX package \cite{cvx,gb08}. During test, we use the label
of its nearest neighbor in the training set as the predicted test
label. This is consistent with the convention in the time series community,
in which they use 1NN as classifier.

\subsection{Effectiveness of local distance metric learning}

First, we fix the alignment, and explore the performances of local
metric learning. Then, we analyze the influence of the preceding alignment
qualities on the performances of subsequent metric learning.

\begin{figure*}
\begin{centering}
\includegraphics[width=0.85\textwidth]{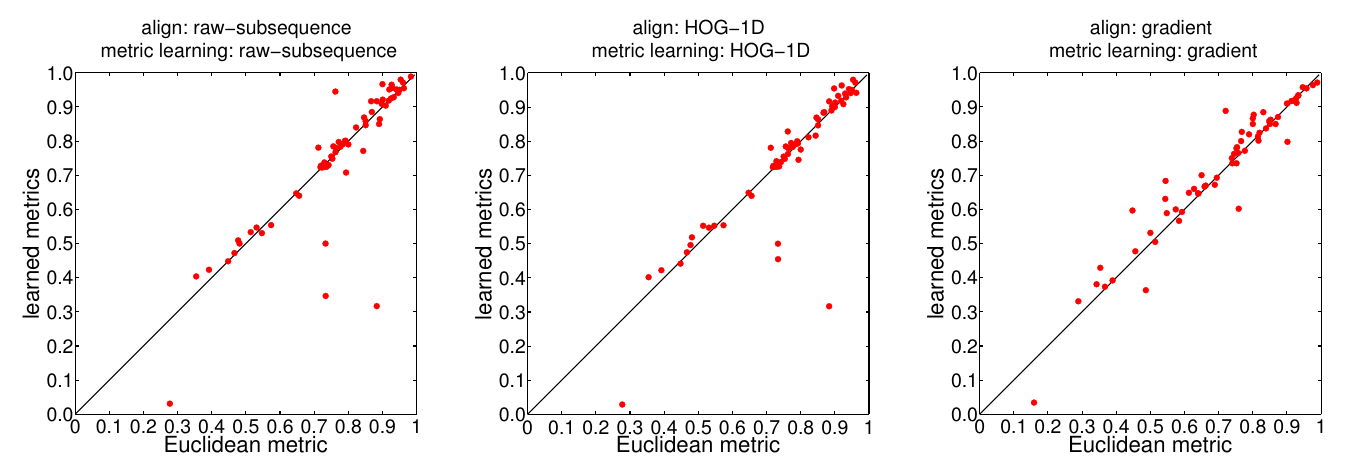} 
\par\end{centering}

\centering{}\caption{\label{fig:metricLearning}Effectiveness of multiple local metrics
learning. Three plots show the comparison between 1NN classifier performances
under the Euclidean metric and the learned metrics. Under all three
descriptors, we obtain significantly improved accuracies, indicating
our proposed multiple local metric learning approach is effective.}
\end{figure*}

We align time series by DTW under three descriptors, derivative, HOG-1D
and raw-subsequence, respectively. Given the computed alignments,
we learn local distance metrics under the same descriptor as used
in the alignment by solving the LP problem (\ref{eq:LMNN-dtw}), and
plot 1NN classification accuracies in Fig. \ref{fig:metricLearning}.
Plots in Fig. \ref{fig:metricLearning} are scatter plots showing
the comparison between 1NN classifier performances under the Euclidean
metric and the learned metrics. Each red dot in the plot indicates
one UCR dataset, whose x-mapping and y-mapping are accuracies under
the Euclidean metric and the learned metrics respectively. By running
the signed rank Wilcoxon test, we obtain p-values 0.038/0.015/0.003
for the descriptor raw-subsequence/HOG-1D/gradient, showing that our
proposed metric learning improve the 1NN classifier significantly
under the confidence level $5\%$.

\begin{figure*}
\begin{centering}
\includegraphics[width=0.85\textwidth]{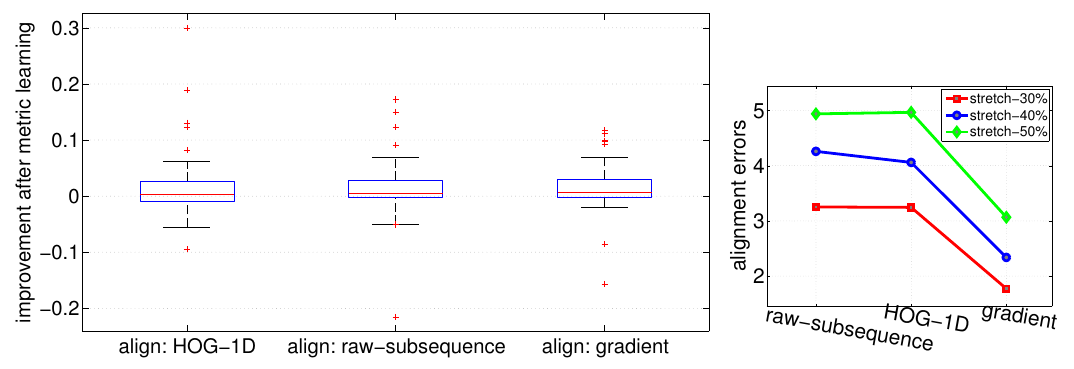} 
\par\end{centering}

\centering{}\caption{\label{fig:Influence-of-alignments}Influence of the alignment qualities
on the metric learning performance. The right plot shows DTW under
different descriptors has different alignment performances. The left
shows that under alignment paths returned by different descriptors,
we execute the subsequent metric learning under the gradient descriptor,
and plot the 1NN performance improvements of the learned metrics to
the Euclidean metric. Even if the preceding alignments have different
qualities, the subsequent metric learning always improves the 1NN
performances significantly (p-values = 0.007/0.029/0.003 under alignments
by the descriptor raw-subsequence/HOG-1D/gradient).}
\end{figure*}

Since the alignment path is the input for the metric learning step,
bad alignments may affect the performance. Nevertheless, we empirically
show this is not the case. We perform metric learning under different
alignments, and evaluate whether significant improvements can be achieved
under all cases. In experiments, we align time series under three
descriptors, and then learn metrics under the gradient descriptor.
We use boxplot to show the performance improvements, compared with
using the default Euclidean metric, in Fig. \ref{fig:Influence-of-alignments}(left).
The blue box has two tails: the lower and upper edge of each blue
box represent 25th and 75th percentiles, with the red line inside
the box marking the median improvement and two tails indicating the
best and worst improvements. Under three different alignments, the
median improvements are all greater than 0 and the majority of improvements
are above 0. By running the signed rank Wilcoxon test between the
1NN performances under the Euclidean metric and the learned metrics,
we obtain p-values 0.007/0.029/0.003 under alignments by the descriptor
raw-subsequence/HOG-1D/gradient. This empirically indicates that the
subsequent metric learning is robust to the preceding alignment qualities.

To show different descriptors do have different alignment qualities,
we could compare DTW alignment paths under different descriptors against
the ground-truth alignments. However, UCR datasets do not have the
ground-truth alignments, here we simulate time series alignment pairs
by manually scaling and stretching time series, and the ground-truth
alignment between the original time series and the stretched one is
known by simulation; then we run DTW alignment under different descriptors,
evaluate the alignment error against the ground truth, and plot the
results in \ref{fig:Influence-of-alignments}(right). It shows different
algorithms do perform differently. We refer the readers to the supplementary
materials for simulation details.

\subsection{Effects of hyper-parameters}

\begin{figure*}
\begin{centering}
\includegraphics[width=0.8\textwidth]{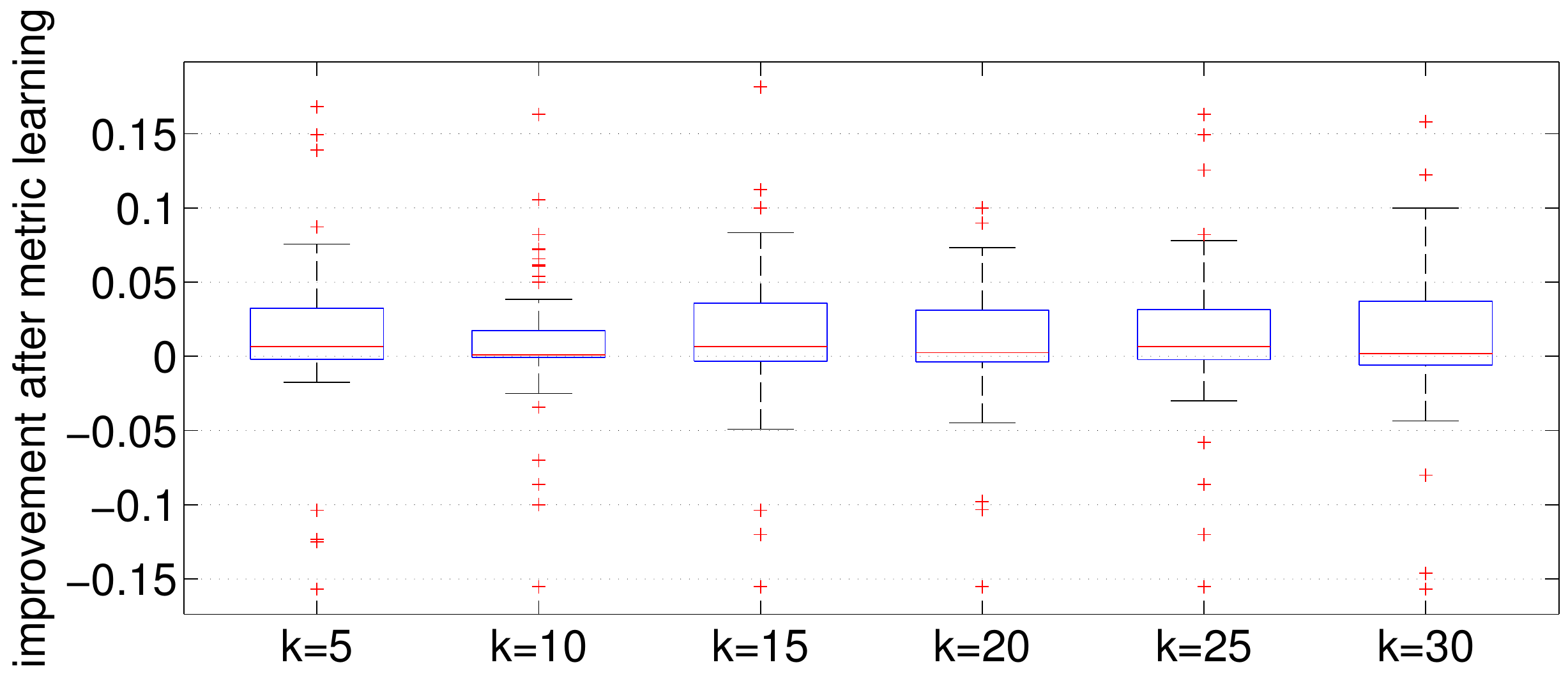} 
\par\end{centering}

\centering{}\caption{\label{fig:Effects-of-different-ks}Effects of different numbers of
descriptor clusters on the metric learning performance. The boxplot
shows the improvements after local metric learning under different
k's. Visually seen, all median improvements are above 0, and the majority
improvements of each boxplot lie above 0 as well. The signed rank
Wilcoxon test shows the significantly improved performances under
all k's.}
\end{figure*}

There is one important hyper-parameter in the metric learning: the
number of clusters of descriptors. In experiments, we align and learn
local metrics both under the gradient descriptor, and during the metric
learning, we set different numbers of descriptor clusters, i.e., $k=\{5,10,15,20,25,30\}$,
learn metrics by solving (\ref{eq:LMNN-dtw}), and plot the 1NN performance
improvements in Fig. \ref{fig:Effects-of-different-ks}. Under different
k's, the majority of the improvements are above 0, and the signed
rank Wilcoxon test returns p-values 0.003/ 0.026/ 0.005/ 0.021/ 0.002/
0.017 under k=5/10/15/20/25/30, showing significant improvements under
varied k's.

\subsection{Comparison with the state of the art algorithm}

As shown in \cite{petitjean2014dynamic,wang2013experimental,bagnall2014experimental,rakthanmanon2012searching},
1NN classifier with the DTW distance as the similarity measure (1NN-DTW)
is very hard to beat. Here we use 1NN-DTW as the baseline, and compare
our algorithms to it. In 1NN-DTW, the alignment is computed by DTW
as well, however, no descriptor is used, i.e., the point to point
distance is directly computed by the squared Euclidean distance between
those two points, instead of by their descriptor distance. The DTW
distance between two aligned sequences is computed as the accumulated
squared Euclidean point-to-point distances with no descriptor used
as well.

\begin{figure*}
\begin{centering}
\includegraphics[width=0.8\textwidth]{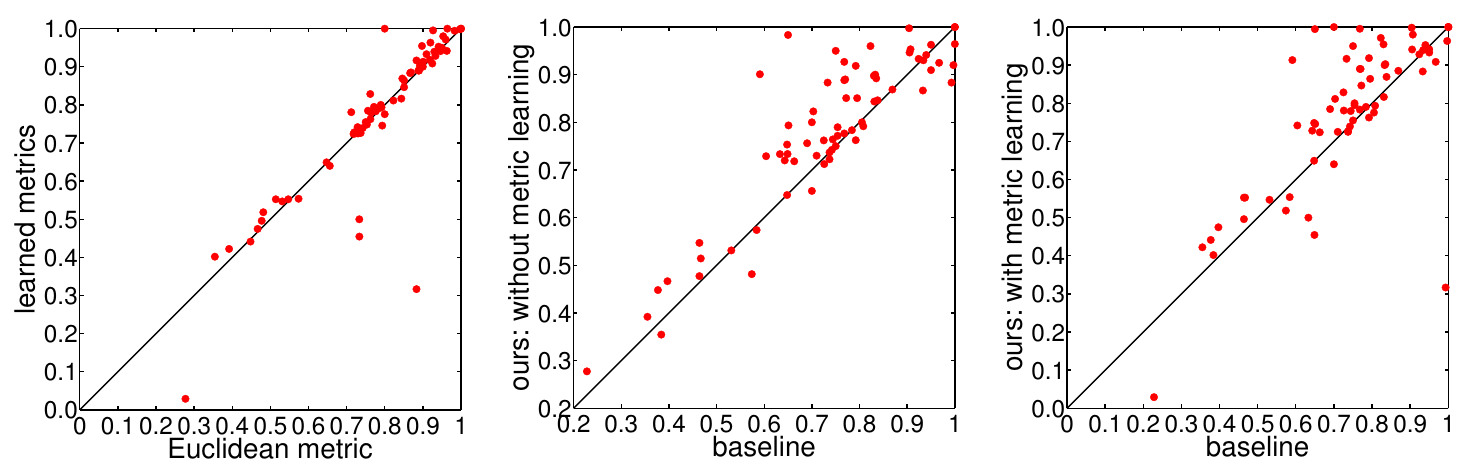} 
\par\end{centering}

\centering{}\caption{\label{fig:Comparison-with-1NN-DTW.}Comparison with 1NN-DTW. Left:
comparison between our algorithm under metric learning and under the
default Euclidean metric; Middle and Right: comparison between our
algorithm under no metric (under metric learning) and the baseline.
The hypothesis test shows our algorithm wins the baseline significantly,
and reaches the state-of-the-art performances on 70 UCR time series
datasets.}
\end{figure*}

In our case, we use the HOG-1D descriptor to align sequences and learn
local metrics. We plot the time series classification performances
in Fig. \ref{fig:Comparison-with-1NN-DTW.}: our algorithm with (without)
metric learning wins/draws/loses the baseline on 48/3/19 (47/3/20)
datasets, and the signed rank Wilcoxon test returns p-values $1.1\cdot10^{-4}(1.8\cdot10^{-5})$,
showing significant accuracy improvement over 1NN-DTW. We document
the classification error rates of three algorithms in the supplementary
materials.

\section{Conclusion and discussion}

In this paper, we propose to learn multiple local Mahalanobis distance
metrics to perform k-nearest neighbor (kNN) classification of temporal
sequences. We showed empirically that the metric learning process
always improves the 1NN time series classification accuracies, disregard
of the qualities of the preceding DTW alignments. Our algorithm wins
the 1NN-DTW algorithm significantly on 70 UCR time series datasets,
and sets up a record for further comparison.

DTW time series classification has two consecutive steps: time series
alignment and then classification. In this paper, the metric learning
happens after the alignment finishes, and information in the metric
learning does not back-prop into the preceding alignment step. An
naive extension is to do alignment and metric learning in an iterative
process. But as we tried, this deteriorated the classification performances.
A future research direction is how to do the alignment and learn metrics
in an integrated fashion.


%

\small

\bibliographystyle{ieee}
\bibliography{metricDTW}


\end{document}